\documentclass{article}

    \PassOptionsToPackage{numbers, compress, sort}{natbib}

\usepackage{bbm}
\usepackage{makecell}
     \usepackage[preprint]{neurips_2022}

\usepackage[utf8]{inputenc} 
\usepackage[T1]{fontenc}    
\usepackage{url}            
\usepackage{booktabs}       
\usepackage{amsfonts}       
\usepackage{nicefrac}       
\usepackage{microtype}      
\usepackage{xcolor}         
\usepackage{graphicx}
\usepackage{amsmath}
\usepackage{booktabs}
\usepackage{amssymb}
\usepackage{wrapfig}
\usepackage{makecell}
\usepackage{multirow}
\usepackage{diagbox}
\usepackage{algorithmicx}
\usepackage{algorithm}

\usepackage{color}
\usepackage{float}
\usepackage{colortbl}
\definecolor{pretty-blue}{RGB}{0, 113, 188}
\definecolor{icmlblue}{rgb}{0,0.08,0.45} 
\definecolor{linecolor1}{gray}{.95} 
\definecolor{linecolor}{gray}{.895} 
\def\eg{{\it{e.g.}}}
\def\etal{{\it{et al.}}}

\usepackage[colorlinks=True,
            linkcolor=red,
            anchorcolor=blue,  
            pagebackref,
            urlcolor=black,
            citecolor=pretty-blue,
            ]{hyperref}

\title{PointDistiller: Structured Knowledge Distillation Towards Efficient and Compact 3D Detection}

\author{
Linfeng Zhang~$^{13*}$\qquad
Runpei Dong~$^2$\thanks{The first two authors have equal contributions. This work is done during internship of L. Zhang in DIDI.}\qquad
Hung-Shuo Tai~$^3$\qquad
Kaisheng Ma~$^1$\thanks{Corresponding author.}\\
Tsinghua University~$^1$\qquad
Xi'an Jiaotong University~$^2$\qquad
DIDI~$^3$ 
}

\begin{document}

\maketitle
\begin{abstract}
    The remarkable breakthroughs in point cloud representation learning have boosted their usage in real-world applications such as self-driving cars and virtual reality. However, these applications usually have an urgent requirement for not only accurate but also efficient 3D object detection. 
    Recently, knowledge distillation has been proposed as an effective model compression technique, which transfers the knowledge from an over-parameterized teacher to a lightweight student and achieves consistent effectiveness in 2D vision. However, due to point clouds' sparsity and irregularity, directly applying previous image-based knowledge distillation methods to point cloud detectors usually leads to unsatisfactory performance.
    To fill the gap, this paper proposes PointDistiller, a structured knowledge distillation framework for point clouds-based 3D detection.
    Concretely, PointDistiller includes \emph{local distillation} which extracts and distills the local geometric structure of point clouds with dynamic graph convolution and \emph{reweighted learning} strategy, which highlights student learning on the crucial points or voxels to improve knowledge distillation efficiency.
    Extensive experiments on both voxels-based and raw points-based detectors have demonstrated the effectiveness of our method over seven previous knowledge distillation methods. 
    For instance, our 4$\times$ compressed PointPillars student achieves 2.8 and 3.4 mAP improvements on BEV and 3D object detection, outperforming its teacher by 0.9 and 1.8 mAP, respectively. Codes have been released at \url{https://github.com/RunpeiDong/PointDistiller} .
    
\end{abstract}
\section{Introduction}
The growth in large-scale lidar datasets~\citep{kitti} and the achievements in end-to-end 3D representation learning~\citep{pointnet,qi2017pointnet++} have boosted the developments of point cloud based segmentation, generation, and detection~\citep{lang2019pointpillars,qi2021offboard}.
As one of the essential tasks of 3D computer vision, 3D object detection plays a fundamental role in real-world applications such as autonomous driving cars~\citep{kitti,DBLP:conf/cvpr/ChenMWLX17,caesar2020nuscenes} and virtual reality~\citep{DBLP:conf/ismar/ParkLW08}. However, recent research has shown a growing discrepancy between cumbersome 3D detectors that achieve state-of-the-art performance and lightweight 3D detectors which are affordable in real-time applications on edge devices.
To address this problem, sufficient model compression techniques have been proposed, such as network pruning~\citep{deepcompression,admm_pruning,pruing_l0,metapruing}, quantization~\citep{data_free_quantization,actication_quantization,dgms}, lightweight model design~\citep{mobilenet3,mobilenetv2,shufflenet}, and knowledge distillation~\citep{distill_hinton}.

\begin{figure}
    \begin{minipage}[t]{0.66\textwidth}

        \centering
    \includegraphics[width=9cm]{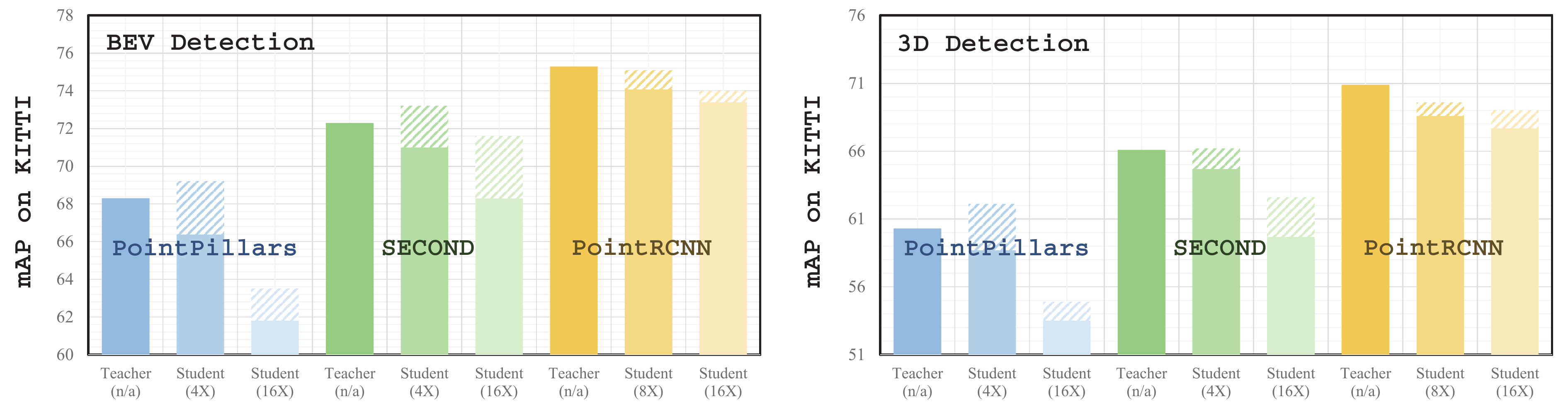}
    \vspace{-0.2cm}
    \caption{Results (mAP of moderate difficulty) of our methods on 4$\times$, 8$\times$, and 16$\times$ compressed students on KITTI. The area of dash lines indicates the benefits of knowledge distillation.}
    \vspace{-0.3cm}
    \label{fig:result}
    \end{minipage}
    \begin{minipage}[t]{0.02\textwidth}
        ~~
    \end{minipage}
    \begin{minipage}[t]{0.29\textwidth}

        \centering
        \includegraphics[width=4cm]{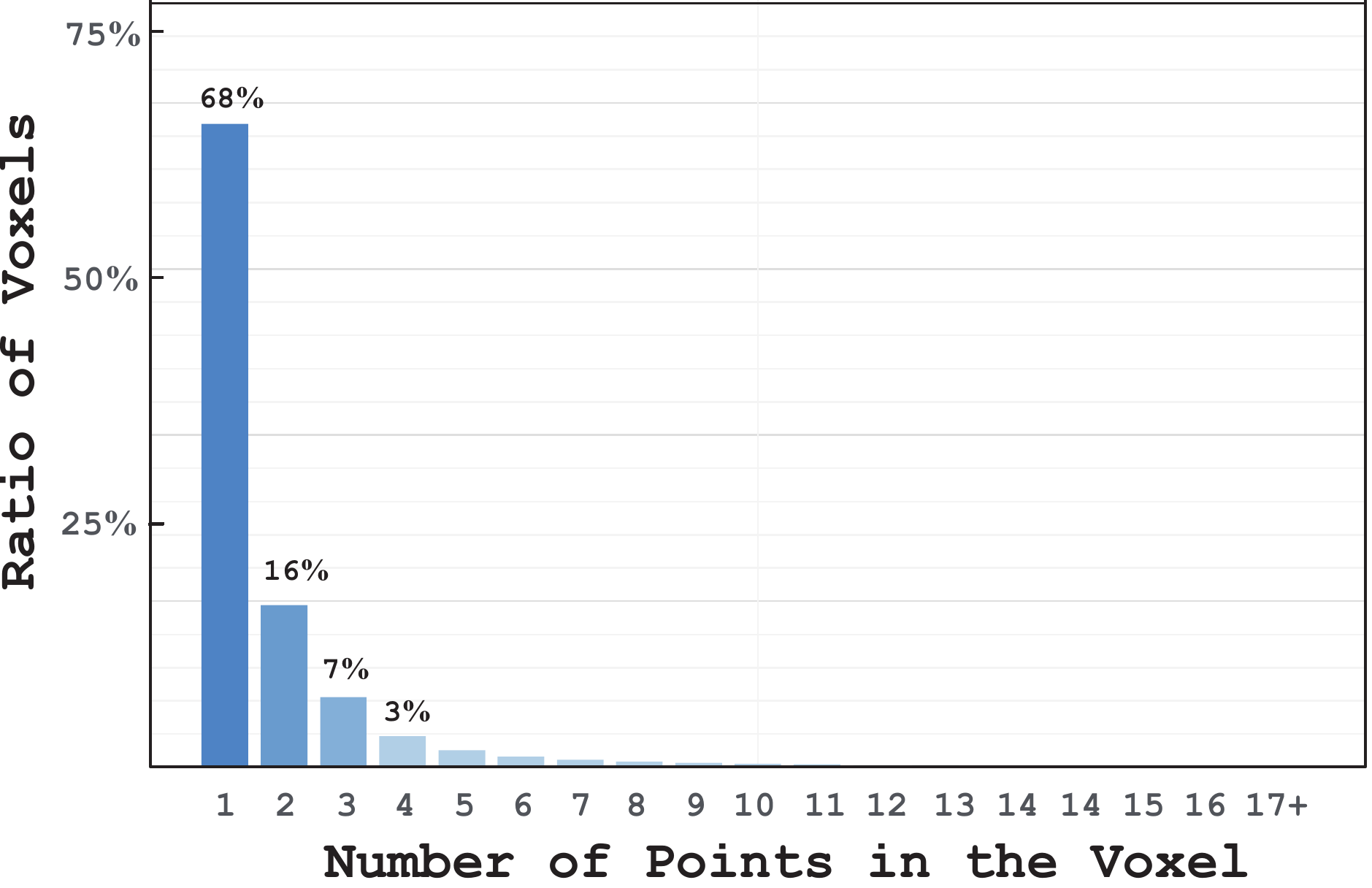}
        \vspace{-0.55cm}
        \caption{Distribution of the voxels with different number of points inside them.}
        \vspace{-0.3cm}
        \label{fig:motivation}
    \end{minipage}

\end{figure}

Knowledge distillation, which aims to improve the performance of a lightweight student model by training it to mimic a pre-trained and over-parameterized teacher model, has evolved into one of the most popular and effective model compression methods in both computer vision and natural language processing~\citep{distill_hinton,fitnets,kd_bert1,kd_bert2}. Sufficient theoretical and empirical results have demonstrated its effectiveness in image-based visual tasks such as image classification~\citep{distill_hinton,fitnets}, semantic segmentation~\citep{structured_kd} and object detection~\citep{kd_detection1,kd_detection2,detectiondistillation,kd_detection4}. 
However, compared with images, point clouds have their properties: (i) Point clouds inherently lack topological information, which makes recovering the local topology information crucial for the visual tasks~\citep{PointConv,PointMLP,DeepGCNs}. (ii) Different from images that have a regular structure, point clouds are irregularly and sparsely distributed in the metric space~\citep{SparseConv,SST}. 

These differences between images and point clouds have hindered the image-based knowledge distillation methods from achieving satisfactory performance on point clouds and also raised the requirement to design specific knowledge distillation methods for point clouds. Recently, a few methods have been proposed to apply knowledge distillation to 3D detection~\citep{cross_modal,sautier2022image,LIGA}. However, most of these methods focus on the choice of student-teacher in a multi-modal setting, \emph{e.g.}, teaching point clouds-based student detectors with an images-based teacher or vice versa, and still ignore the peculiar properties of point clouds. To address this problem, we propose a structured knowledge distillation framework named PointDistiller, which involves \emph{local distillation} to distill teacher knowledge in the local geometric structure of point clouds, and \emph{reweighted learning} strategy to handle the sparsity of point clouds by highlighting student learning on the relatively more crucial voxels.

\emph{\textbf{Local Distillation~}} 
Sufficient recent studies show that capturing and making usage of the semantic information in the local geometric structure of point clouds have a crucial impact on point cloud representation learning~\citep{qi2017pointnet++,graphconv5}. 
Hence, instead of directly distilling the backbone feature of teacher detectors to student detectors, we propose local distillation, which firstly clusters the local neighboring voxels or points with KNN (K-Nearest Neighbours), then encodes the semantic information in local geometric structure with dynamic graph convolutional layers~\citep{graphconv5}, and finally distill them from teachers to students. Hence, the student detectors can inherit the teacher's ability to understand point clouds' local geometric information and achieve better detection performance.

\emph{\textbf{Reweighted Learning Strategy~}} 
One of the mainstream methods for processing point clouds is to convert them into volumetric voxels and then encode them as regular data.
However, due to the sparsity and the noise in point clouds, most of these voxels contain only a single point. For instance, as shown in Figure~\ref{fig:motivation}, on the KITTI dataset, around 68\% voxels in point clouds contain only one point, which has a high probability of being a noise point. Hence, the representative features in these single-point voxels have relatively lower importance in knowledge distillation compared with the voxels which contain multiple points. Motivated by this observation, we propose a reweighted learning strategy, which highlights student learning on the voxels with multiple points by giving them larger learning weights. Besides, the similar idea can also be easily extended to raw points-based detectors to highlight knowledge distillation on the points  which have a more considerable influence on the prediction of the teacher detector.

Extensive experiments on both voxels-based and raw-points based detectors have been conducted to demonstrate the effectiveness of our method over the previous seven knowledge distillation methods. As shown in Figure~\ref{fig:result}, on PointPillars and SECOND detectors, our method leads to 4$\times$ compression and 0.9$\sim$1.8 mAP improvements at the same time. On PointRCNN, our method leads to 8$\times$ compression with only 0.2 BEV mAP drop. 
Our main contributions be summarized as follows.
\begin{itemize}
    \item
    We propose \emph{local distillation}, which firstly encodes the local geometric structure of point clouds with dynamic graph convolution and then distills them from teachers to students.
    \item We propose \emph{reweighted learning strategy} to handle the sparsity and noise in point clouds. It highlights student learning on the voxels, which have more points inside them, by giving them higher learning weights in knowledge distillation.
    \item Extensive experiments on both voxels-based and raw points-based detectors have been conducted to demonstrate the performance of our method over seven previous methods. Besides, we have released our codes to promote future research.
\end{itemize}

\section{Related Work}

\vspace{-0.05cm}
\subsection{Knowledge Distillation}  
\vspace{-0.05cm}
The idea of training a small model with a large pre-trained model is firstly proposed by
Buciluǎ~\emph{et~al.} for ensemble model compression~\citep{model_compression}.
Then, with the excellent breakthroughs of deep learning, Hinton~\emph{et al.} propose the concept of knowledge distillation which strives to compress an over-parameterized teacher model by transferring its knowledge to a lightweight student model~\citep{distill_hinton}. Early knowledge distillation methods usually train the students to mimic the predicted categorical probability distribution of teachers~\citep{distill_hinton,deepmutuallearning}. Then, extensive methods have been proposed to learn teacher knowledge in the backbone features~\citep{fitnets} or its variants, such as attention~\citep{attentiondistillation,detectiondistillation}, relation~\citep{relational_kd,relational_kd2,kd_cc,spkd_gan}, task-oriented information~\citep{tofd} and so on. Following its success in classification, abundant works have applied knowledge distillation to object detection~\citep{detectiondistillation,kd_detection1,kd_detection2,kd_detection3}, segmentation~\citep{structured_kd}, image generation~\citep{wkd,spkd_gan,omgd,revisit_discriminator,gan_compress,teacher_do_more_gankd}, pre-trained language models~\citep{kd_bert1,kd_bert2}, semi-supervised learning~\citep{kd_semi,kd_semi2} and lead to consistent effectiveness.

\vspace{-0.05cm}

\paragraph{Knowledge Distillation on Object Detection}
Recently, designing specific knowledge distillation methods to improve the efficiency and accuracy of object detection has become a rising and popular topic. Chen~\emph{et al.} first propose to apply the naive prediction and feature-based knowledge distillation methods to object detection~\citep{kd_detection1}. Then, Wang~\emph{et al.} show that the imbalance between foreground objects and background objects hinders knowledge distillation from achieving better performance in object detection~\citep{kd_detection3}. To address this problem, abundant knowledge distillation methods have tried to find the to-be-distilled regions based on the ground-truth~\citep{kd_detection3}, detection results~\citep{dai2021general}, spatial attention~\citep{detectiondistillation}, query-based attention~\citep{kang2021instance} and gradients~\citep{DBLP:conf/cvpr/Guo00W0X021}. 
Moreover, recent methods have also been proposed to distill the pixel-level and object-level relation from teachers to students~\citep{detectiondistillation,structuredkd,dai2021general}. 
Besides knowledge distillation for 2D detection, some cross-modal knowledge distillation have been introduced to transfer knowledge from RGB-based teacher detectors to lidar-based student detectors or vice versa~\citep{cross_modal,sautier2022image,chong2022monodistill,LIGA}. However, most of these methods focus on the choice of students and teachers in a multi-modal framework, while the design of specific knowledge distillation optimization methods on point clouds based pure 3D detection has not been well-explored.

\vspace{-0.2cm}

\subsection{3D Object Detection on Point Clouds}

\vspace{-0.2cm}

The rapid development of deep learning has firstly boosted the research in 2D object detection and then recently raised the research trend in point clouds-based 3D object detection. PointNet~\citep{pointnet} is firstly proposed to extract the feature of points with multi-layer perception in an end-to-end manner. Then, 
PointNet++ is further proposed to capture the local structures in a hierarchical fashion with density adaptive sampling and grouping~\citep{qi2017pointnet++}.
Zhou~\emph{et al.} propose VoxelNet, a single-stage detector that divides a point cloud into equally spaced 3D voxels and processes them with voxel feature encoding layers~\citep{zhou2018voxelnet}.
Then, SECOND is proposed to improve VoxelNet with sparse convolutional layers and focal loss~\citep{yan2018second}. PointPillars is proposed to divide point clouds into several pillars and then convert them into a pseudo image, which can be further processed with 2D convolutional layers~\citep{lang2019pointpillars,paigwar2021frustum}.
Shi~\emph{et al.} propose PointRCNN, a two-stage detection method that firstly generates bottom-up 3D proposals based on the raw point clouds and then refines them to obtain the final detection results~\citep{shi2019pointrcnn}. Afterward, Fast Point R-CNN and PV-RCNN are proposed to utilize both voxel representation and raw point clouds to exploit their respective advantages~\citep{chen2019fast,PVRCNN}. Recently, Qi~\etal{} propose to perform offboard 3D detection with point cloud sequences, which is able to make use of the temporal points and achieve comparable performance with human labels~\citep{qi2021offboard}.
The graph convolutional neural network is another rising star in point cloud detection~\citep{graphconv4,graphconv5}. Lin~\emph{et~al.} propose 3D-GCN to avoid the shift and scale changes in point clouds~\citep{graphconv1}. Zhou~\emph{et al.} propose adaptive graph convolution, which generates adaptive kernels according to the learned features~\citep{graphconv2}. 

\vspace{-0.2cm}

\paragraph{Efficient 3D Object Detectors}
Unfortunately, the significant 3D detection performance usually comes at the expense of high computational and storage costs, making them unaffordable in real-time applications such as self-driving cars. To address this issue, recent research attention has been paid to designing efficient 3D detectors. Tang~\emph{et al.} propose to apply neural architecture search to 3D detection by using sparse point-voxel convolution~\citep{tang2020searching}. Li~\emph{et al.} propose Lidar-RCNN, which resorts to a point-based approach and remedies the problem of uncorrected proposal sizes~\citep{li2021lidar}. Liu~\emph{et al.} propose voxel-point cnn to represent the 3D input data in points while performing the convolutions in voxels to reduce the memory accessing consumption~\citep{liu2019point}. Recently, Li~\emph{et al.} propose to improve the efficiency of graph convolution for point clouds by simplified KNN search and graph shuffling~\citep{graphconv3}.

\vspace{-0.1cm}
\section{Methodology\label{sec:method}}
\vspace{-0.1cm}
\subsection{Preliminaries}
Given a set of point clouds $\mathcal{X}=\{x_1, x_2, ..., x_n\}$ and the corresponding label set $\mathcal{Y}=\{y_1, y_2, ..., y_m\}$, the object detector can be formulated as $\mathcal{F}=f \circ g$, where $f$ is the feature encoding layer to extract representation features from inputs and $g$ is the detection head for prediction. Then, the representation feature on the sample $x$ can be written as $f(x)\in \mathbb{R}^{n\times C}$, where $n$ indicates the number of voxels for voxels-based detectors or the number of points for raw points-based detectors. $C$ indicates the number of channels. 
Besides, for voxels-based detectors, we define $v_{ij}(x)=1$ if the $j$-th point of $x$ belongs to the $i$-voxel else 0. Then, the number of points in the $i$-th voxel can be denoted as $\sum_j v_{ij}(x)$. 
Usually, knowledge distillation involves a to-be-trained student detector and a pre-trained teacher detector, and we distinguish them with scripts $\mathcal{S}$ and $\mathcal{T}$, respectively. 
\begin{figure}
    \centering
    \includegraphics[width=\linewidth]{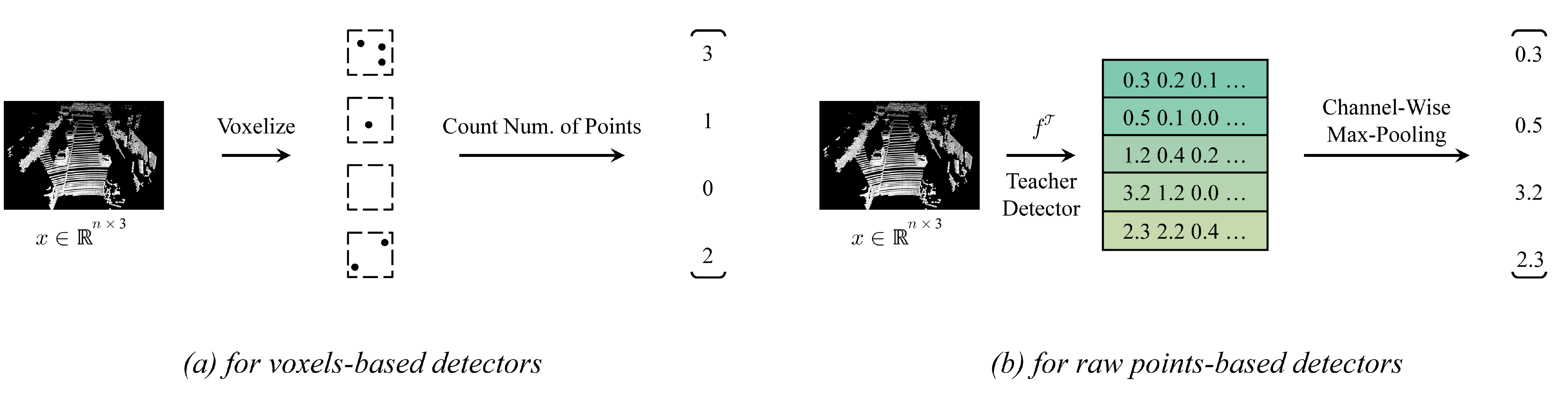}
    \caption{The computation of the importance score for voxels-based detectors and raw-points-based detectors. The importance scores are later utilized to determine which voxel or point is utilized for distillation and how they contribute to the distillation loss.}\label{fig:score_comput}
  
\end{figure}
\begin{figure}
    \centering
    \includegraphics[width=\linewidth]{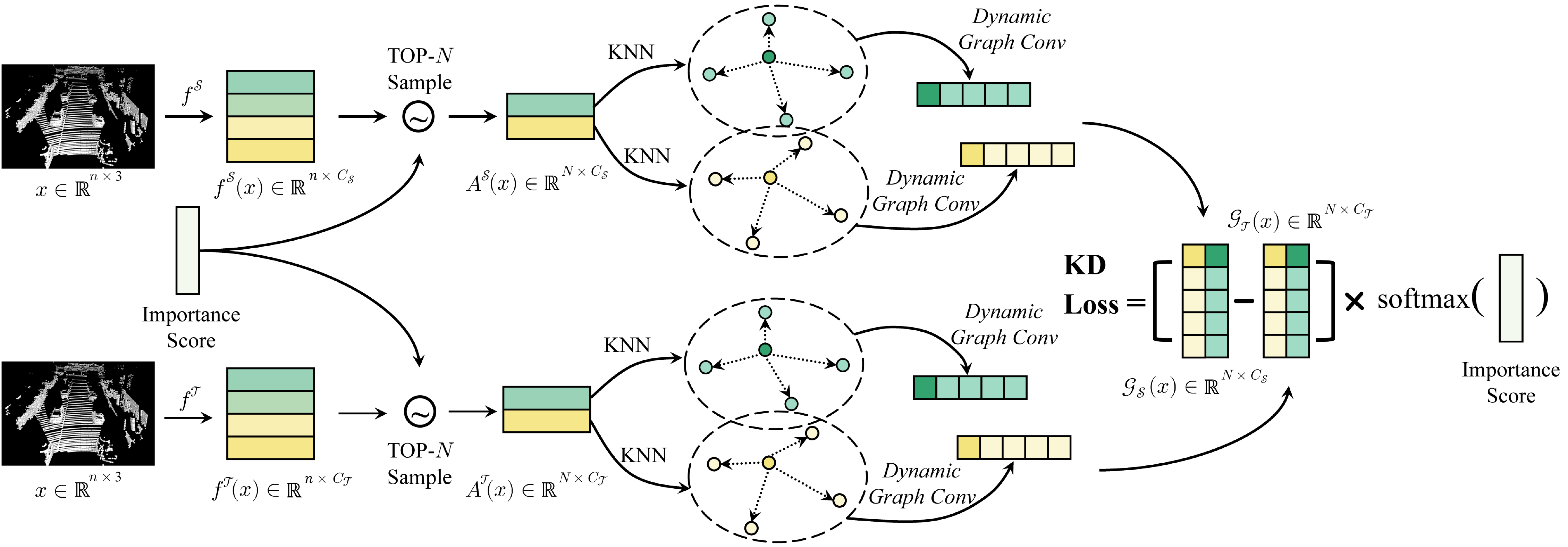}

    \caption{The details of our method. 
    $f^{\mathcal{T}}$ and $f^{\mathcal{S}}$: the feature encoding layers in the teacher and student detectors.
    $A^{\mathcal{T}}$ and $A^{\mathcal{S}}$: features of the sampled to-be-distilled voxels or points with top-$N$ largest importance score. 
    $C_{\mathcal{T}}$ and $C_{\mathcal{S}}$: the number of channels for features of the teacher and the student detectors.
    $\mathcal{G}_{\mathcal{T}}$ and $\mathcal{G}_{\mathcal{S}}$: the graph features of the teacher and student detectors.
    Based on the pre-defined importance score, our method samples the relatively more crucial $N$ voxels or points from the whole point cloud, extracts their local geometric structure of them with dynamic graph convolution, and then distills them in a reweighted manner. 
    Please refer to Section~\ref{sec:method} for more details. }\label{fig:method}
  
\end{figure}

\subsection{Our Method}
\paragraph{Sampling Top-$N$ To-be-distilled Voxels (Points)} As discussed in previous sections, since the point clouds are overwhelmingly sparse while the voxels are usually equally spaced, most of the voxels only contain very few and even single point. Thus, these single-point voxels have much less value to be learned by students in knowledge distillation.
Even in raw points-based detectors, there usually exist some points which are relatively more crucial and some points which are not meaningful (\eg{}, the noise points). Thus, instead of distilling all the voxels or points in point clouds, we propose to distill the voxels or points which are more valuable for knowledge distillation. Concretely, for voxels-based detectors, we define the importance score of $i$-th voxel as $\sum_j v_{ij}(x)$, which indicates the number of points inside it. For point-based detectors, motivated by previous works which localized the crucial pixels in images with attention,
we define the importance score for $i$-th point as its permutation-invariant maximal value along the channel dimension, which can be formulated as $\text{max}\left(f(x)[i]\right)$. Based on the importance score, we can sample the top-$N$ significant voxels or points for knowledge distillation based on the importance score computed from $f^{\mathcal{T}}(x)$. 
For simplicity in writing, we denote the selected student and teacher features in top-$N$ important voxels or points as $A^{\mathcal{T}}(x) \in \mathbb{R}^{N\times C_\mathcal{T}}$ and $A^{\mathcal{S}}(x) \in \mathbb{R}^{N\times C_\mathcal{S}}$, respectively, where $C_\mathcal{S}$ and $C_\mathcal{T}$ indicate the number of channels in student and teacher features.

\paragraph{Extracting Local Geometric Information} As pointed out by abundant previous works, the local geometric information has a crucial influence on the performance of point cloud detectors~\citep{qi2017pointnet++,graphconv5}. Thus, instead of directly distilling the representative feature, we propose \emph{local distillation} which extracts the local geometric information of point clouds with dynamic graph convolution layers and distills it to the student detector. Concretely, denoting $z_i=A(x)[i]$ as the feature of the $i$-th to-be-distilled voxel or point, we can build a graph based on this voxel or point and its $K$ neighboring voxels or points clustered by KNN (K-Nearest Neighbours). By denoting the features of $z_i$ and its $K-1$ neighbours as $z_{i,1}$ and $\mathcal{N}_i = \{z_{i,2}, z_{i,3},..., z_{i,K}\}$ respectively, motivated by previous methods~\citep{pointnet,graphconv5},
we firstly update the feature of each voxel (or point) in this graph by concatenating them with the global centroid voxel (or point) feature $z_{i,1}$, which can be formulated as $\hat{z}_{i,j} = \text{cat}\big([z_{i,1}, z_{i,j}]\big)$ for all $z_{i,j}\in\mathcal{N}_i$. Then, we apply a dynamic graph convolution as the aggregation operation upon them, which can be formulated as $\mathcal{G}_i = \gamma(\hat{z}_{i,1}, ..., \hat{z}_{i,K})$, where $\gamma$ is the aggregation operator. Following previous graph-based point cloud networks, we set $\gamma$ as a nonlinear layer with ReLU activation and batch normalization.
Then the training objective of local distillation can be formulated as 
\begin{equation}
    \mathop{\arg\min}\limits_{\mathbf{\theta}_\mathcal{S}, \mathbf{\theta}_{\gamma}}
    \mathbb{E}_{~x} \left[
    \frac{1}{N} \mathop{\sum}_{i=1}^N  \left\| \mathcal{G}_i^\mathcal{S}(x) -  \mathcal{G}_i^\mathcal{T}(x)    
    \right\|\right],
\end{equation}
where $\theta_{\mathcal{S}}$ indicates the parameters of student encoding layer $f^{\mathcal{S}}$. $\theta_{\gamma} = [\theta_{\gamma^{\mathcal{S}}}, \theta_{\gamma^{\mathcal{T}}}]$ indicates the parameters of dynamic graph convolution layers for the student and teacher detectors. Note that these layers are trained with the student detector simultaneously and can be discarded during inference. 

\paragraph{Reweighting Knowledge Distillation Loss}
Usually, compared with the teacher detector, the student detector has much fewer parameters, implying inferior learning capacity. Thus, it is challenging for the student detector to inherit teacher knowledge in all points or voxels. As discussed above, different voxels and points in point cloud object detection have different values in knowledge distillation. Thus, we propose to reweight the learning weight of each voxel or point based on the importance score introduced in previous paragraphs. Denote the learning weight for the $N$ to-be-distilled as $\phi \in \mathbb{R}^{N}$. Similar with the importance score defined during sampling, we define the learning weight of each graph as the maximal value on the corresponding features after a softmax function, which can be formulated as $\phi = \text{softmax}\left(\text{max}(G^\mathcal{T}(x)) / \tau\right)$, where $\tau$ is the temperature hyper-parameter in softmax function. For voxels-based methods, we define $\phi$ as the number of points in the voxel after a softmax function, which can be formulated as $\phi_i = \text{softmax}\left(\sum_j v_{i,j}
 / \tau\right)$. 
Then, with the reweighting strategy, the training objective of knowledge distillation can be formulated as 
\begin{equation}
\label{loss}
    \mathop{\arg\min}_{\mathbf{\theta}_\mathcal{S}, \mathbf{\theta}_{\gamma}}
    \mathbb{E}_{~x} \left[
    \frac{1}{N} 
    \mathop{\sum}_{i=1}^N  \phi_i \cdot \left\| \mathcal{G}_i^\mathcal{S}(x) -  \mathcal{G}_i^\mathcal{T}(x)    
    \right\|\right].
\end{equation}
As shown in the above loss function, with a higher $\phi_i$, the knowledge distillation loss between student and teacher features at the $i$-th graph will have a more extensive influence on the overall loss, and thus student learning on the $i$-th graph can be highlighted. As a result, the proposed reweighting strategy allows the student detector to pay more attention to learning teacher knowledge in the relatively more crucial voxel graphs (point graphs). Moreover, Equation~\ref{loss} also implies that our method is a feature-based knowledge distillation method that is not correlated with the architecture of detectors and the label set $\mathcal{Y}$. Hence, it can be directly added to the origin training loss of all kinds of 3D object detectors for model compression. 

\vspace{-0.25cm}
\section{Experiment}
\vspace{-0.1cm}
\subsection{Experiment Setting}\label{sec:exp_set}
We have evaluated our method in both voxels-based object detector including PointPillars~\citep{lang2019pointpillars} and SECOND~\citep{yan2018second}, and the raw points based object detector including PointRCNN~\citep{shi2019pointrcnn}.
Most experiments are conducted on KITTI~\citep{kitti} and nuScenes~\citep{caesar2020nuscenes}, which consist of samples that have both lidar point clouds and images. Our models are trained with only the lidar point clouds. For KITTI, we report the average precision calculated by 40 sampling recall positions for BEV (Bird's Eye View) object detection and 3D object detection on the \textit{validation} split. Following the typical protocol, the IoU threshold is set as 0.7 for class Car and 0.5 for class Pedestrians and Cyclists. 
We have mainly compared our methods with seven previous knowledge distillation methods, including methods proposed by Remero~\emph{et al.}~\citep{fitnets}, Zagoruko~\emph{et al.}~\citep{attentiondistillation}, Tung~\emph{et al.}~\citep{spkd_gan}, Heo~\emph{et al.}~\citep{kd_comprehensive}, Zheng~\emph{et al.}~\citep{DBLP:conf/cvpr/ZhengTJF21}, Tian~\etal{}~\citep{kd_crd},
 and Zhang~\etal{}~\citep{detectiondistillation}.
All the experiments are conducted with mmdetection3d~\citep{mmdet3d2020} and PyTorch~\citep{pytorch}. We keep the training and evaluation settings in mmdetection3d as default. The teacher model is the origin model before compression. The student model shares the same architecture and depth as its teacher but with fewer channels. 
Following previous works, the average precision of three difficulties and the three categories are reported as the performance metrics~\citep{kitti}. 
Please refer to our codes in the supplementary material for more details.

\begin{table*}[t!]
\caption{\label{tab:BEV} Experimental results of our method for BEV (Bird-Eye-View) object detection. \textbf{F} and \textbf{P} indicate the number of float operations (/G) and
parameters (/M) of the detector, respectively. \textbf{mAP} indicates the mean average precision of moderate difficulty. \textbf{KD} indicates whether our method is utilized. The reported result in the first line of each detector is the performance of the teacher detector.
}
\vspace{-0.15cm}
\begin{center}
    \footnotesize
  \setlength{\tabcolsep}{0.90mm}{ \begin{tabular}{lllccccccccccccc}
  \toprule[1.25pt]
   \multirow{2}{*}[-0.5ex]{Model}&\multirow{2}{*}[-0.5ex]{{F}}&\multirow{2}{*}[-0.5ex]{P}&\multirow{2}{*}[-0.5ex]{KD}&\multicolumn{3}{c}{Car}&&\multicolumn{3}{c}{Pedestrians}&&\multicolumn{3}{c}{Cyclists}&\multirow{2}{*}[-0.5ex]{mAP}\\
   \cmidrule(){5-7}\cmidrule(){9-11}\cmidrule(){13-15} 
   &&&&Easy&Moderate&Hard&&Easy&Moderate&Hard&&Easy&Moderate&Hard&\\
     \midrule[0.65pt]
    \multirow{5}{*}[-1.0ex]{PointPillars}
    &34.3&4.8&$\times$&94.3&88.1&83.6 && 57.9&51.8&47.6 && 86.5&65.0&61.1 & 68.3\\
    \cmidrule(){2-16}
    &9.0&1.3&$\times$&92.4&88.2&83.6 && 53.0&47.9&44.1 && 81.8&63.1&59.0 & 66.4\\ 
    &\cellcolor{linecolor}9.0&\cellcolor{linecolor}1.3&\cellcolor{linecolor}$\checkmark$&\cellcolor{linecolor}\textbf{93.1}&\cellcolor{linecolor}\textbf{89.0}&\cellcolor{linecolor}\textbf{86.3} &\cellcolor{linecolor}&\cellcolor{linecolor}\textbf{59.8}&\cellcolor{linecolor}\textbf{52.8}&\cellcolor{linecolor}\textbf{48.2} &\cellcolor{linecolor}&\cellcolor{linecolor}\textbf{83.8}&\cellcolor{linecolor}\textbf{65.8}&\cellcolor{linecolor}\textbf{62.0} &\cellcolor{linecolor}\textbf{69.2}\\
    \cmidrule(){2-16}
    &2.5&0.3&$\times$& 91.3&84.8&\textbf{82.2} && 50.1&44.4&41.6 && 74.2&56.1&52.5& 61.8\\
    \rowcolor{linecolor}\cellcolor{white}&2.5&0.3&$\checkmark$&\textbf{92.5}&\textbf{85.2}&81.9 && \textbf{50.8}&\textbf{45.8}&\textbf{42.5} && \textbf{77.2}&\textbf{59.5}&\textbf{55.6}& \textbf{63.5}\\
     \midrule[0.65pt]
    \multirow{5}{*}[-1.0ex]{SECOND}
    &69.8&5.3&$\times$&93.1&88.9&85.9 && 64.9&58.1&51.9 && 84.3&69.9&65.7& 72.3\\
    \cmidrule(){2-16}
    &17.8&1.4&$\times$&93.1&86.6&85.7 && 64.7&57.8&52.8 && 84.1&68.5&64.5 & 71.0\\
    &\cellcolor{linecolor}17.8&\cellcolor{linecolor}1.4&\cellcolor{linecolor}$\checkmark$&\cellcolor{linecolor}\textbf{93.2}&\cellcolor{linecolor}\textbf{88.6}&\cellcolor{linecolor}\textbf{86.0} &\cellcolor{linecolor}&\cellcolor{linecolor}\textbf{65.1}&\cellcolor{linecolor}\textbf{58.1}&\cellcolor{linecolor}\textbf{53.1} &\cellcolor{linecolor}&\cellcolor{linecolor}\textbf{87.4}&\cellcolor{linecolor}\textbf{72.9}&\cellcolor{linecolor}\textbf{68.5} &\cellcolor{linecolor}\textbf{73.2}\\
    \cmidrule(){2-16}
    &4.6&0.4&$\times$&95.0&86.2&83.3 && 61.6&54.9&49.2 && 80.9&63.6&59.6 & 68.3\\
    \rowcolor{linecolor}\cellcolor{white}
    &4.6&0.4&$\checkmark$&\textbf{95.4}&\textbf{88.3}&\textbf{83.7} && \textbf{64.5}&\textbf{57.6}&\textbf{52.2} && \textbf{85.2}&\textbf{68.8}&\textbf{64.4} & \textbf{71.6}\\
    \midrule[0.65pt]
   \multirow{5}{*}[-1.0ex]{PointRCNN}
   &104.9&4.1&$\times$&95.0&86.7&84.3 && 69.8&64.5&58.1 && 92.8&74.6&70.4 & 75.3\\
   \cmidrule(){2-16}
   &13.7&0.5&$\times$&\textbf{93.5}&\textbf{85.9}&\textbf{83.5} && 71.6&65.4&59.1 && 91.1&71.0&67.2 & 74.1\\
   &\cellcolor{linecolor}13.7&\cellcolor{linecolor}0.5&\cellcolor{linecolor}$\checkmark$&\cellcolor{linecolor}93.3&\cellcolor{linecolor}85.7&\cellcolor{linecolor}\textbf{83.5}&\cellcolor{linecolor}&\cellcolor{linecolor}\textbf{74.0}&\cellcolor{linecolor}\textbf{67.2}&\cellcolor{linecolor}\textbf{60.5}&\cellcolor{linecolor}&\cellcolor{linecolor}\textbf{94.6}&\cellcolor{linecolor}\textbf{72.3}&\cellcolor{linecolor}\textbf{67.9}&\cellcolor{linecolor}\textbf{75.1}\\
    \cmidrule(){2-16}
    &7.1&0.3&$\times$&\textbf{95.8}&\textbf{85.4}&\textbf{81.7} && \textbf{72.9}&\textbf{65.5}&\textbf{58.6} && 91.8&69.3&65.9&73.4\\
    \rowcolor{linecolor}\cellcolor{white}
    &7.1&0.3&$\checkmark$&95.2&84.3&\textbf{81.7} && 72.6&64.8&57.7&&\textbf{92.6}&\textbf{72.9}&\textbf{68.5}&\textbf{74.0}\\
    \bottomrule[1.25pt]\\
\end{tabular}}
\vspace{-0.6cm}
\end{center}
\end{table*}   
\begin{table*}[t!]
\vspace{-0.6cm}
\caption{\label{tab:3d}
Experimental results of our method for 3D object detection. \textbf{F} and \textbf{P} indicate the number of float operations (/G) and
parameters (/M) of the detector, respectively. \textbf{mAP} indicates the mean average precision of moderate difficulty. \textbf{KD} indicates whether our method is utilized. 
The reported result in the first line of each detector is the performance of the teacher detector.
}
\vspace{-0.15cm}
\begin{center}
    \footnotesize
  \setlength{\tabcolsep}{0.90mm}{ \begin{tabular}{lllccccccccccccc}
  \toprule[1.25pt]
   \multirow{2}{*}[-0.5ex]{Model}&\multirow{2}{*}[-0.5ex]{{F }}&\multirow{2}{*}[-0.5ex]{P}&\multirow{2}{*}[-0.5ex]{KD}&\multicolumn{3}{c}{Car}&&\multicolumn{3}{c}{Pedestrians}&&\multicolumn{3}{c}{Cyclists}&\multirow{2}{*}[-0.5ex]{mAP}\\
   \cmidrule(){5-7}\cmidrule(){9-11}\cmidrule(){13-15} 
   &&&&Easy&Moderate&Hard&&Easy&Moderate&Hard&&Easy&Moderate&Hard&\\
     \midrule[0.65pt] 
    \multirow{5}{*}[-1.0ex]{PointPillars}
    &34.3&4.8&$\times$&87.3&75.9&71.1 && 52.0&45.9&41.4 && 78.6&59.2&55.8 & 60.3\\
    \cmidrule(){2-16}
    &9.0&1.3&$\times$&87.4&75.9&71.0 && 48.2&43.0&38.7 && 74.1&57.2&53.3 & 58.7\\
    &\cellcolor{linecolor}9.0&\cellcolor{linecolor}1.3&\cellcolor{linecolor}$\checkmark$&\cellcolor{linecolor}\textbf{88.1}&\cellcolor{linecolor}\textbf{76.9}&\cellcolor{linecolor}\textbf{73.8} &\cellcolor{linecolor}&\cellcolor{linecolor}\textbf{54.6}&\cellcolor{linecolor}\textbf{47.5}&\cellcolor{linecolor}\textbf{42.3} &\cellcolor{linecolor}&\cellcolor{linecolor}\textbf{80.3}&\cellcolor{linecolor}\textbf{62.0}&\cellcolor{linecolor}\textbf{58.8} &\cellcolor{linecolor}\textbf{62.1}\\
    \cmidrule(){2-16}
    &2.5&0.3&$\times$&83.1&\textbf{69.8}&\textbf{65.4} && 44.0&38.7&35.3 && 70.9&52.1&48.7 & 53.5\\
    \rowcolor{linecolor}\cellcolor{white}
    &2.5&0.3&$\checkmark$&\textbf{83.7}&\textbf{69.8}&65.3 && \textbf{45.3}&\textbf{40.3}&\textbf{36.5} && \textbf{72.7}&\textbf{54.7}&\textbf{51.1} & \textbf{54.9}\\
     \midrule[0.65pt] 
    \multirow{5}{*}[-1.0ex]{SECOND}
    &69.8&5.3&$\times$&88.6&79.3&75.7 && 60.1&53.2&47.0 && 79.8&65.7&61.6 & 66.1\\
    \cmidrule(){2-16}
    &17.8&1.4&$\times$&\textbf{89.2}&\textbf{77.4}&\textbf{74.0} && 58.8&51.3&45.5 && 80.5&65.4&61.3 & 64.7\\
    &\cellcolor{linecolor}17.8&\cellcolor{linecolor}1.4&\cellcolor{linecolor}$\checkmark$&\cellcolor{linecolor}88.9&\cellcolor{linecolor}76.9&\cellcolor{linecolor}73.6 &\cellcolor{linecolor}&\cellcolor{linecolor}\textbf{60.0}&\cellcolor{linecolor}\textbf{53.0}&\cellcolor{linecolor}\textbf{47.4} &\cellcolor{linecolor}&\cellcolor{linecolor}\textbf{83.2}&\cellcolor{linecolor}\textbf{68.6}&\cellcolor{linecolor}\textbf{64.2} &\cellcolor{linecolor}\textbf{66.2}\\
    \cmidrule(){2-16}
    &4.6&0.4&$\times$&86.3&72.6&66.0 && 53.6&47.8&41.8 && 76.7&58.7&55.1 & 59.7\\
    \rowcolor{linecolor}\cellcolor{white}
    &4.6&0.4&$\checkmark$&\textbf{87.0}&\textbf{73.3}&\textbf{68.1} && \textbf{57.0}&\textbf{51.0}&\textbf{45.4} && \textbf{81.0}&\textbf{63.5}&\textbf{59.3} & \textbf{62.6}\\
    \midrule[0.65pt]
   \multirow{5}{*}[-1.0ex]{PointRCNN}
    &104.9&4.1&$\times$&92.1&80.1&77.4 && 66.8&60.3&54.3 && 92.1&72.3&67.8 & 70.9\\
    \cmidrule(){2-16}
    &13.7&0.5&$\times$&89.8&\textbf{76.8}&72.7 && 67.9&60.9&54.0 && 88.1&68.0&64.4 & 68.6\\
    &\cellcolor{linecolor}13.7&\cellcolor{linecolor}0.5&\cellcolor{linecolor}$\checkmark$&\cellcolor{linecolor}\textbf{91.4}&\cellcolor{linecolor}75.6&\cellcolor{linecolor}\textbf{72.9} &\cellcolor{linecolor}&\cellcolor{linecolor}\textbf{70.1}&\cellcolor{linecolor}\textbf{63.5}&\cellcolor{linecolor}\textbf{56.1}&\cellcolor{linecolor}&\cellcolor{linecolor}\textbf{92.0}&\cellcolor{linecolor}\textbf{69.8}&\cellcolor{linecolor}\textbf{65.4} &\cellcolor{linecolor}\textbf{69.6}\\
    \cmidrule(){2-16}
    &7.1&0.3&$\times$&\textbf{89.8}&75.3&70.7&&68.7&60.7&53.4&&\textbf{91.1}&67.2&63.9&67.7\\
    \rowcolor{linecolor}\cellcolor{white}
    &7.1&0.3&$\checkmark$&89.6&\textbf{75.6}&\textbf{72.6} && \textbf{69.4}&\textbf{61.0}&\textbf{53.5} && 91.0&\textbf{70.2}&\textbf{65.5} & \textbf{69.0}\\
    \bottomrule[1.25pt]
\end{tabular}}
\vspace{-0.1cm}
\end{center}
\end{table*}

\vspace{-0.1cm}
\subsection{Experimental Results}
 \vspace{-0.1cm}
 
 Table~\ref{tab:BEV} and Table~\ref{tab:3d} show the performance of detectors trained with and without our method for BEV detection and 3D detection, respectively. It is observed that: (i) Significant average precision improvements on all kinds of detectors and all compression ratios for both BEV and 3D detection. On average, 2.4 and 1.0 moderate mAP improvements can be observed for the voxel and raw points-based detectors, respectively. On BEV and 3D detection, 1.9 and 1.9 moderate mAP improvements can be obtained, respectively.
(ii) On the BEV detection of PointPillars and SECOND detectors, the 4$\times$ compressed and accelerated students trained with our method outperform their teachers by 0.9 and 0.9 mAP, respectively.
On the 3D detection of PointPillars and SECOND detectors, the 4$\times$ compressed and accelerated students trained with our method outperform their teachers by 1.8 and 0.1 mAP, respectively.
(iii) Consistent average precision boosts can be observed in detection results of all difficulties. For instance, on BEV detection of PointPillars students, 2.4, 2.3, and 2.3 mAP improvements can be observed for easy, moderate, and hard difficulties, respectively. These observations demonstrate that our method can successfully transfer teacher knowledge to the student detectors. 
(iv) Consistent average precision boosts can be observed in detection results of all categories. For instance, on moderate BEV detection of PointPillars students, 0.6, 3.2 and 3.1 mAP improvements can be obtained on cars, pedestrians and cyclists, respectively. (v) On PointRCNN, on average 1.3 and 1.2 moderate mAP improvements can be observed on BEV and 3D detection, respectively, indicating that our method is also effective for raw points-based detectors.
In summary, these experiment results demonstrate that our method can successfully transfer the knowledge from teacher detectors to student detectors and lead to significant and consistent performance boosts.

\vspace{-0.2cm}
\paragraph{Comparison with Other KD Methods}
\vspace{-0.2cm}
Comparison between our method and previous knowledge distillation methods is shown in Table~\ref{tab:compare}. It is observed that: (i) Our method outperforms the previous methods by a clear margin. On BEV and 3D detection, our method outperforms the second-best knowledge distillation method by 1.5 and 1.9 moderate mAP, respectively. (ii) Our method achieves the best performance for all categories of all difficulties. (iii) Besides, our method is the only knowledge distillation method that enables the student detector to outperform its teacher detector.

\begin{table*}
\caption{\label{tab:compare} Comparison between our method and previous knowledge distillation methods on PointPillars. The teacher and the student detectors have 34.3 and 9.0 GFLOPs, respectively. \textbf{mAP} indicates the mean average precision of moderate difficulty.
}
\vspace{-0.1cm}
\begin{center}
    \footnotesize
  \setlength{\tabcolsep}{0.9mm}{
  \resizebox{\columnwidth}{!}{
  \begin{tabular}{llcccccccccccccc}
  \toprule[1.25pt]
   \multirow{2}{*}[-0.5ex]{Task}&\multirow{2}{*}[-0.5ex]{Method}&\multicolumn{3}{c}{Car}&&\multicolumn{3}{c}{Pedestrians}&&\multicolumn{3}{c}{Cyclists}&&\multirow{2}{*}[-0.5ex]{mAP}\\
   \cmidrule(){3-5}\cmidrule(){7-9}\cmidrule(){11-13} 
   &&Easy&Moderate&Hard&&Easy&Moderate&Hard&&Easy&Moderate&Hard&\\
     \midrule[0.65pt]
   \multirow{10}{*}[-0.5ex]{BEV}&Teacher w/o KD&94.3&88.1&83.6 && 57.9&51.8&47.6 && 86.5&65.0&61.1 && 68.3\\
  \cmidrule(){2-15}
    &Student w/o KD&92.4&88.2&83.6 && 53.0&47.9&44.1 && 81.8&63.1&59.0 && 66.4\\ 
     &+ Romero~\emph{et al.}~\citep{fitnets}&91.5&85.6&83.1&&57.5&51.0&46.3&&82.8&65.1&61.1&&67.2\\
     &+ Zagoruyko~\emph{et al.}~\citep{attentiondistillation}&92.6&88.0&83.6&&56.7&50.9&47.3&&81.4&64.4&60.5&&67.7\\
     &+ Zheng~\emph{et al.}~\citep{DBLP:conf/cvpr/ZhengTJF21}&92.7&87.9&83.2&&57.7&51.0&46.8&&78.1&61.8&57.9&&66.9\\
     &+ Tung~\emph{et al.}~\citep{relational_kd2}&92.8&88.0&83.3&&54.5&48.7&45.2&&84.2&64.3&60.7&&67.0\\
     &+ Tian~\emph{et al.}~\citep{kd_crd}&92.7&87.8&83.2&&56.6&50.4&46.8&&80.3&61.9&57.9&&66.7\\
     &+ Heo~\emph{et al.}~\citep{kd_comprehensive}&92.6&87.9&83.5&&57.6&51.0&46.8&&78.1&61.8&57.8&&66.9\\
     &+ Zhang~\emph{et al.}~\citep{detectiondistillation}&92.3&85.7&83.0&&59.7&52.0&47.6&&71.0&64.3&60.5&&67.5
     \\
     \rowcolor{linecolor}\cellcolor{white}
     &+ Ours&\textbf{93.1}&\textbf{89.0}&\textbf{86.3} && \textbf{59.8}&\textbf{52.8}&\textbf{48.2} && \textbf{83.8}&\textbf{65.8}&\textbf{62.0} && \textbf{69.2}\\
   \midrule[0.65pt]
   \multirow{10}{*}[-0.5ex]{3D}&Teacher w/o KD&87.3&75.9&71.1 && 52.0&45.9&41.4 && 78.6&59.2&55.8 && 60.3\\
  \cmidrule(){2-15}
    &Student w/o KD&87.4&75.9&71.0 && 48.2&43.0&38.7 && 74.1&57.2&53.3 && 58.7\\
     &+ Romero~\emph{et al.}~\citep{fitnets}&84.9&73.4&70.6&&50.9&44.2&39.3&&75.9&58.5&54.6&&58.7\\
     &+ Zagoruyko~\emph{et al.}~\citep{attentiondistillation}&87.6&75.7&71.4&&51.0&44.8&40.7&&74.4&57.8&54.2&&59.5\\
     &+ Zheng~\emph{et al.}~\citep{DBLP:conf/cvpr/ZhengTJF21}&87.3&75.5&71.5&&52.6&45.6&40.8&&74.9&58.6&54.9&&59.9\\
     &+ Tung~\emph{et al.}~\citep{relational_kd2}&87.5&76.0&71.3&&50.1&43.3&39.2&&79.2&59.5&55.3&&59.6\\
     &+ Tian~\emph{et al.}~\citep{kd_crd}&85.6&74.2&71.0&&49.5&43.5&39.0&&76.4&58.4&54.7&&58.7\\
     &+ Heo~\emph{et al.}~\citep{kd_comprehensive}&87.7&76.1&71.7&&52.6&45.6&40.8&&74.9&58.6&54.9&&60.1\\
     &+ Zhang~\emph{et al.}~\citep{detectiondistillation}&87.5&75.8&71.6&&53.4&45.8&40.9&&76.1&59.0&55.2&&60.2
     \\
    \rowcolor{linecolor}\cellcolor{white}
     &+ Ours&\textbf{88.1}&\textbf{76.9}&\textbf{73.8} && \textbf{54.6}&\textbf{47.5}&\textbf{42.3} && \textbf{80.3}&\textbf{62.0}&\textbf{58.8} && \textbf{62.1}\\
    \bottomrule[1.25pt]
\end{tabular}
}
}
\vspace{-0.3cm}
\end{center}
\end{table*}

\begin{figure}
    \begin{minipage}[t]{0.5\textwidth}

        \centering
    \includegraphics[width=\textwidth]{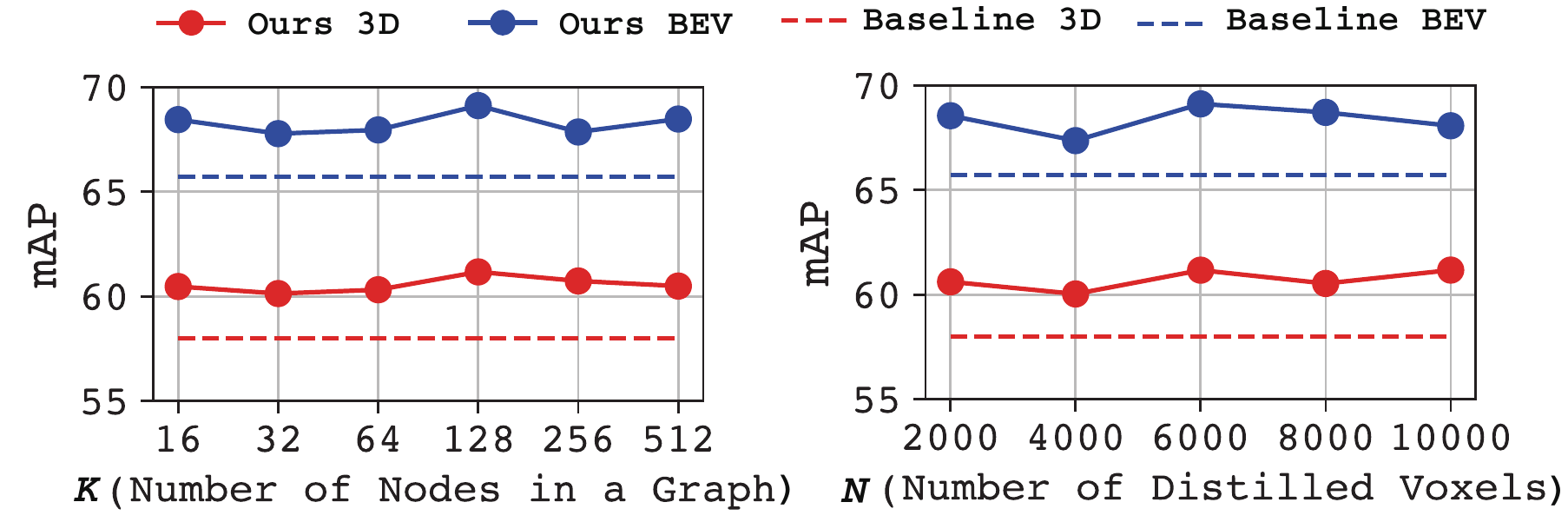}
    
    \caption{Hyper-parameters sensitivity study on KITTI with 4$\times$ compressed PointPillars detctors. mAP is measured on the moderate difficulty.}
    \label{fig:sensivity}
    \end{minipage}
    \begin{minipage}[t]{0.02\textwidth}
    \end{minipage}
    \begin{minipage}[t]{0.48\textwidth}

        \centering
        \includegraphics[width=\textwidth]{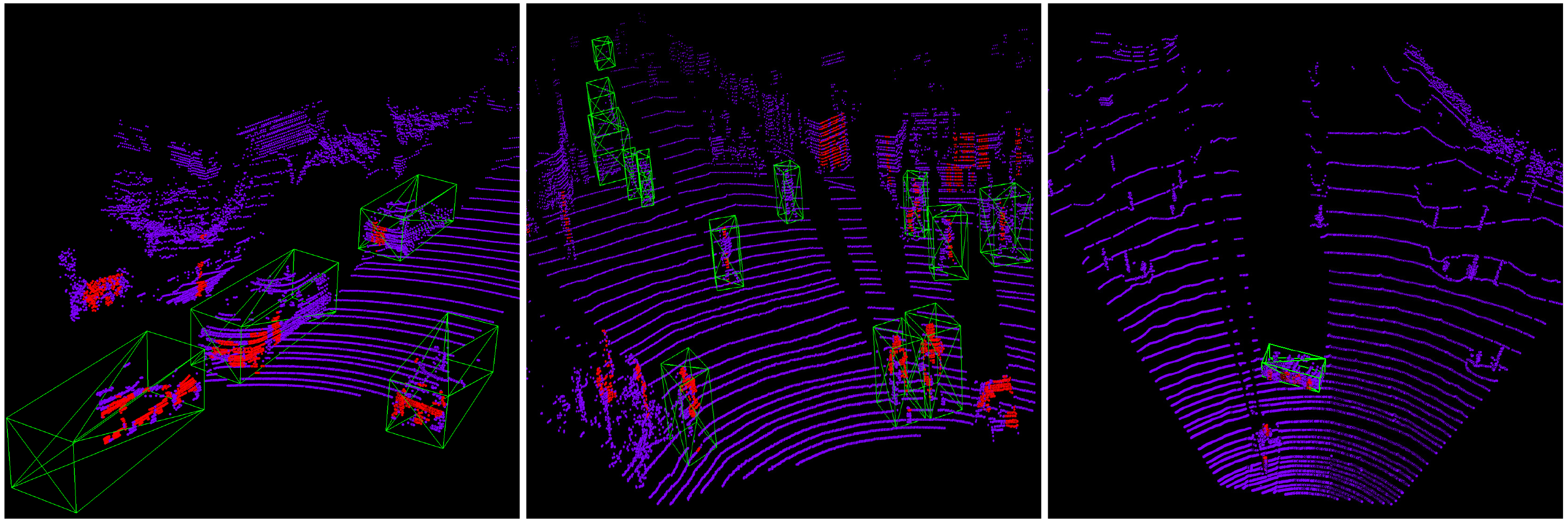}
        
        \caption{Visualization on the importance scores for PointPillars. Red points indicate the voxels with high importance scores. }
        \label{fig:score}
    \end{minipage}
\vspace{-0.2cm}
\end{figure}
\begin{table}[h]
    \caption{Experimental results on nuScenes dataset with PointPillars. \textbf{CR} indicates the compression ratio. \textbf{KD} indicates whether knowledge distillation is utilized. A higher mAP and NDS, and a lower mATE, mASE, mAOE, mAVE and mAAE indicate better performance.\label{tab:nuscene}}
    \begin{center}
  \setlength{\tabcolsep}{1.5mm}{
  \footnotesize
  \begin{tabular}{lccccccccccccccccc}
    \toprule
        Model & CR &KD &mAP($\uparrow$)&NDS($\uparrow$)&mATE($\downarrow$)&mASE($\downarrow$)&mAOE($\downarrow$)&mAVE($\downarrow$)&mAAE($\downarrow$)\\
         \midrule
         \multirow{4}{*}[-0.5ex]{PointPillars}&\multirow{2}{*}{2}&$\times$&36.0&50.5&\textbf{44.8}&28.3&51.2&32.9&18.0\\
         &&\cellcolor{linecolor}$\checkmark$&\cellcolor{linecolor}\textbf{36.7}&\cellcolor{linecolor}\textbf{51.0}&\cellcolor{linecolor}44.9&\cellcolor{linecolor}\textbf{28.1}&\cellcolor{linecolor}\textbf{51.0}&\cellcolor{linecolor}\textbf{31.7}&\cellcolor{linecolor}\textbf{17.4}\\
         \cmidrule{2-10}
         &\multirow{2}{*}{4}&$\times$&32.2&47.3&46.7&\textbf{28.4}&60.1&35.9&\textbf{17.2}\\
         &&\cellcolor{linecolor}$\checkmark$&\cellcolor{linecolor}\textbf{32.8}&\cellcolor{linecolor}\textbf{48.6}&\cellcolor{linecolor}\textbf{45.2}&\cellcolor{linecolor}\textbf{28.4}&\cellcolor{linecolor}\textbf{52.3}&\cellcolor{linecolor}\textbf{35.1}&\cellcolor{linecolor}17.3\\
         \bottomrule 
    \end{tabular}}
    \end{center}
\end{table}

\vspace{-0.1cm}
\paragraph{Experiments on nuScenes}
\vspace{-0.1cm}
Experiments of 2$\times$ and 4$\times$ compressed PointPillars on nuScenes are shown in Table~\ref{tab:nuscene}. It is observed that our method leads to 0.65 and 0.5 improvements on mAP and NDS on average, respectively, indicating that our method is also effective on the large-scale dataset.

\section{Discussion}
\vspace{-0.1cm}
\subsection{Ablation Study and Sensitivity Study}
\vspace{-0.1cm}
\paragraph{Ablation Study} The proposed PointDistiller is mainly composed of two components, including the reweighted learning strategy (\texttt{RL}) and local distillation (\texttt{LD}). Ablation studies with 4$\times$ compressed PointPillars students on KITTI are shown in Table~\ref{tab:ablation}. It is observed that: (i) 2.0 and 1.9 mAP improvements can be obtained by only using the reweighted learning strategy to distill the backbone features on BEV detection and 3D detection, respectively.
(ii) 2.3 and 2.5 mAP boosts can be gained by using local distillation without reweighted learning on BEV detection and 3D detection, respectively. 
(iii) By combining the two methods together, 0.5 and 0.9 further mAP improvements can be achieved on BEV detection and 3D detection, respectively. 
These observations indicate that each module in PointDistiller has its individual effectiveness and their merits are orthogonal. Besides, they also implies that the proposed local distillation and reweighted learning may be combined with other knowledge distillation methods to achieve better performance.
\vspace{-0.3cm}
\paragraph{Sensitivity Study}
Our method mainly introduces two hyper-parameters, $K$, and $N$, which indicate the number of nodes in a graph for local distillation, and the number of to-be-distilled voxels (points) respectively. A hyper-parameter sensitivity study on the two hyper-parameters is shown in Figure~\ref{fig:sensivity}. It is observed that our method with different hyper-parameter values consistently outperforms the baseline by a large margin, indicating our method is not sensitive to hyper-parameters.

\vspace{-0.15cm}
\subsection{Visualization Analysis}

\paragraph{Visualization on Importance Score} In the reweighted learning strategy, the importance scores of each voxel or point are utilized to determine whether it should be distilled. Visualization of the importance scores in PointPillars is shown in Figure~\ref{fig:score}. It is observed that they successfully localize the foreground objects (\emph{e.g.,} cars and pedestrians) and the hard-negative objects (\emph{e.g.,} walls).

\begin{figure}
    \centering
    \vspace{-0.5cm}
    \includegraphics[width=14cm]{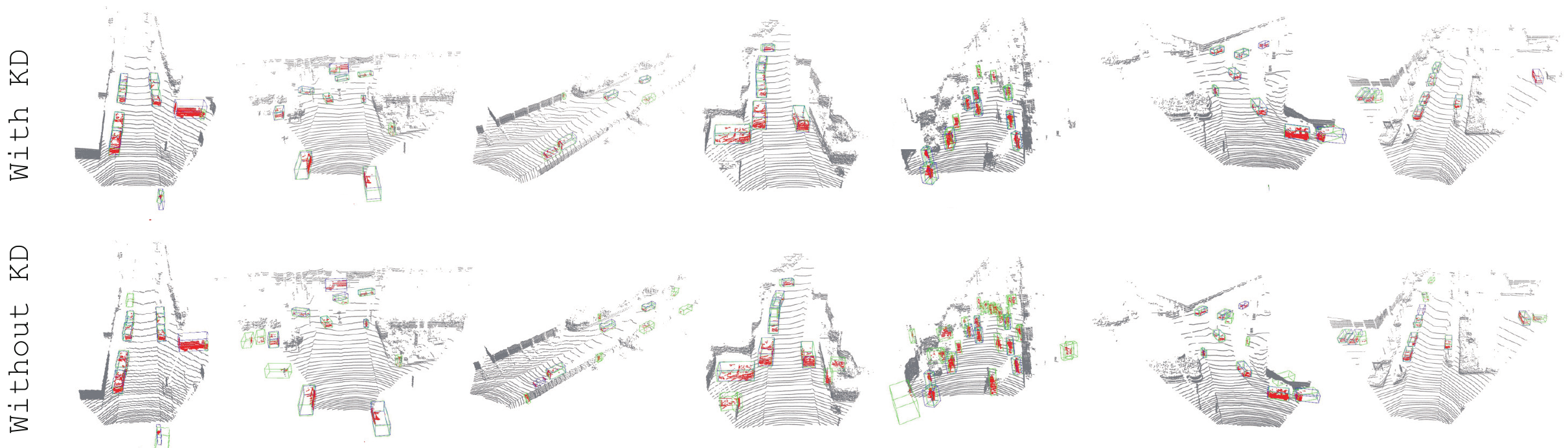}
    \caption{Qualitative comparison between the detection results of students trained with and without knowledge distillation. Green boxes and blue boxes indicate the bounding boxes from the prediction results and the ground-truths. Red points are the points insides the prediction bounding boxes.}
    \vspace{-0.2cm}
    \label{fig:visual}
\end{figure}

\vspace{-0.35cm}
\paragraph{Visualization on Detection Results} In this subsection, we have visualized the detection results of the student model trained with and without our method for comparison. Note that both student models are 4$\times$ compressed PointPillars trained on KITTI. 
The green and blue boxes indicate the boxes of the model prediction and the ground truth. 
As shown in Figure~\ref{fig:visual}, the student model without knowledge distillation tends to have much more false-positive (FP) predictions. 
In contrast, this excessive FP problem is alleviated in the student trained with our method. This observation is consistent with our experimental results that the distilled PointPillars has 3.4 mAP improvements.

\begin{table*}
\caption{\label{tab:ablation} Ablation study on 4$\times$ compressed PointPillars students. \texttt{LD} and \texttt{RL} indicates local distillation and the reweighted learning strategy, respectively. mAP is measured on the moderate difficulty.
}
\begin{center}
    \footnotesize
  \setlength{\tabcolsep}{0.90mm}{ 
  \begin{tabular}{llcccccccccccccc}
  \toprule[1.25pt]
   \multirow{2}{*}[-0.5ex]{Model }&\multirow{2}{*}[-0.5ex]{Task }&\multirow{2}{*}[-0.5ex]{\texttt{LD}}&\multirow{2}{*}[-0.5ex]{\texttt{RL}
   }&\multicolumn{3}{c}{Car}&&\multicolumn{3}{c}{Pedestrians}&&\multicolumn{3}{c}{Cyclists}&\multirow{2}{*}[-0.5ex]{mAP}\\
   \cmidrule(){5-7}\cmidrule(){9-11}\cmidrule(){13-15} 
   &&&&Easy&Moderate&Hard&&Easy&Moderate&Hard&&Easy&Moderate&Hard&\\
     \midrule[0.65pt]
    \multirow{8}{*}[-0.5ex]{PointPillars}
    &\multirow{4}{*}{BEV}&$\times$&$\times$&92.4&88.2&83.6 && 53.0&47.9&44.1 && 81.8&63.1&59.0 & 66.4\\
    &&$\checkmark$&$\times$&92.7&88.2&83.7&&58.2&51.0&47.0&&\textbf{84.3}&66.9&\textbf{63.1}&68.7\\
    &&$\times$&$\checkmark$&\textbf{93.1}&88.5&85.7&&55.6&49.6&45.7&&84.2&\textbf{67.3}&62.9&68.4\\ 
    &&$\checkmark$&$\checkmark$&\textbf{93.1}&\textbf{89.0}&\textbf{86.3} && \textbf{59.8}&\textbf{52.8}&\textbf{48.2} && 83.8&65.8&62.0 & \textbf{69.2}\\
     \cmidrule(){2-16} 
    &\multirow{4}{*}{3D}&$\times$&$\times$&87.4&75.9&71.0 && 48.2&43.0&38.7 && 74.1&57.2&53.3 & 58.7\\
    &&$\checkmark$&$\times$&87.6&76.0&71.5&&52.6&45.9&40.7&&79.8&61.6&58.0&61.2\\
    &&$\times$&$\checkmark$&87.8&76.5&72.0&&49.4&43.7&39.4&&78.7&61.5&57.5&60.6\\
    &&$\checkmark$&$\checkmark$&\textbf{88.1}&\textbf{76.9}&\textbf{73.8} && \textbf{54.6}&\textbf{47.5}&\textbf{42.3} && \textbf{80.3}&\textbf{62.0}&\textbf{58.8} & \textbf{62.1}\\
    \bottomrule[1.25pt]
\end{tabular}}
\vspace{-0.1cm}
\end{center}
\end{table*}

\vspace{-0.15cm}
\section{Conclusion}
\vspace{-0.15cm}
This paper proposes a structured knowledge distillation framework named PointDistiller for point clouds-based object detection. It is composed of \emph{local distillation} to first encode the semantic information in local geometric structure in point clouds and distill it to students, and \emph{reweighted learning} to handle the sparsity and noise in point clouds by assigning different learning weights to different points and voxels.
Extensive experiments on both voxels-based detectors and raw points-based detectors have demonstrated the superiority over seven previous knowledge distillation methods.
Our ablation study has shown the individual effectiveness of each module in PointDistiller. 
Besides, the visualization results demonstrate
that PointDistiller can significantly improve detection performance by reducing false-positive predictions, 
and the importance score is able to reveal the more significant voxels.
To the best of our knowledge, this work initiates the first step to exploring KD for efficient point clouds-based 3D object detection, and we hope this could spur future research.

{
\small
\bibliographystyle{icml2022}
\small
\bibliography{bibs} 
}

\end{document}